\documentclass[a4paper,conference]{IEEEtran}
\usepackage{cite}
\usepackage{amsmath,amssymb,amsfonts}
\usepackage{algorithmic}
\usepackage{graphicx}
\usepackage{textcomp}
\usepackage{url}
\usepackage{xcolor}
\usepackage{times}
\usepackage{epsfig}
\usepackage{amssymb}
\usepackage{fixltx2e}
\usepackage{listings}
\usepackage{booktabs} 
\usepackage{multirow}
\usepackage{gensymb}
\usepackage{mathtools}
\usepackage{adjustbox}
\usepackage{framed}
\usepackage{latexsym}
\newcommand{\etal}{\textit{et al}.}

\usepackage{arydshln}
\usepackage{xcolor}
\definecolor{DarkGreen}{rgb}{0.1,0.5,0.9} 

\makeatletter
\AtBeginDocument{\def\@citecolor{DarkGreen}}
\makeatother

\usepackage[colorlinks,citecolor=DarkGreen,urlcolor=DarkGreen,bookmarks=false,hypertexnames=true]{hyperref} 

\begin{document}
%
\title{ACR Loss: Adaptive Coordinate-based Regression Loss for Face Alignment}

\author{\IEEEauthorblockN{Ali Pourramezan Fard}
\IEEEauthorblockA{School of Electrical and Computer Engineering\\
University of Denver, CO, USA \\
Email: ali.pourramezanfard@du.edu}
\and
\IEEEauthorblockN{Mohammad H. Mahoor}
\IEEEauthorblockA{School of Electrical and Computer Engineering\\
University of Denver, CO, USA \\
Email: mmahoor@du.edu}
}
\maketitle

\begin{abstract}
Although deep neural networks have achieved reasonable accuracy in solving face alignment, it is still a challenging task, specifically when we deal with facial images, under occlusion, or extreme head poses. Heatmap-based Regression (HBR) and Coordinate-based Regression (CBR) are among the two mainly used methods for face alignment. CBR methods require less computer memory, though their performance is less than HBR methods.
In this paper, we propose an Adaptive Coordinate-based Regression (ACR) loss to improve the accuracy of CBR for face alignment. Inspired by the Active Shape Model (ASM), we generate Smooth-Face objects, a set of facial landmark points with less variations compared to the ground truth landmark points. We then introduce a method to estimate the level of difficulty in predicting each landmark point for the network by comparing the distribution of the ground truth landmark points and the corresponding Smooth-Face objects. Our proposed ACR Loss can adaptively modify its curvature and the influence of the loss based on the difficulty level of predicting each landmark point in a face. Accordingly, the ACR Loss guides the network toward challenging points than easier points, which improves the accuracy of the face alignment task. Our extensive evaluation shows the capabilities of the proposed ACR Loss in predicting facial landmark points in various facial images.

We evaluated our ACR Loss using MobileNetV2, EfficientNet-B0, and EfficientNet-B3 on widely used 300W, and COFW datasets and showed that the performance of face alignment using the ACR Loss is much better than the widely-used L2 loss. Moreover, on the COFW dataset, we achieved state-of-the-art accuracy. In addition, on 300W the ACR Loss performance is comparable to the state-of-the-art methods. We also compared the performance of MobileNetV2 trained using the ACR Loss with the lightweight state-of-the-art methods, and we achieved the best accuracy, highlighting the effectiveness of our ACR Loss for face alignment specifically for the lightweight models. 

The code is available on \href{https://github.com/aliprf/ACR-Loss}{Github}

\end{abstract}

$\\$

%
\IEEEpeerreviewmaketitle

\section{Introduction}\label{sec:intro}
Face alignment, often known as facial landmark points detection, is the process of detecting and estimating the location of predefined key-points in facial images. It is a vital step for many computer vision tasks such as face recognition~\cite{adcore9727163, soltanpour2017survey, juhong2017face, ozseven2017face}, head pose estimation~\cite{xia2019head, fard2021asmnet}, facial expression recognition~\cite{sun2014deep, choi2020facial, kalapala2020facial, liu2021facial}, 3D face reconstruction~\cite{liu2016joint, jiang20183d, afzal20203d} Despite recent progress in developing algorithms for face alignment, this task has remained challenging especially for real-world applications, where we deal with face occlusion, scene illumination, and extreme pose variations. 

Deep face alignment can be categorized into two methods: Coordinate-based Regression (CBR)~\cite{feng2018wing, zhang2019facial, sadiq2019facial, yan2020fine, zhang2020phased, fard2021asmnet}, and Heatmap-based Regression (HBR)~\cite{wu2018look, ning2020cpu, wang2019adaptive, liu2019attention, wu2021design}. Although the HBR methods can achieve better accuracy, they are less efficient than CBR methods. To be more specific, while the output of the HBR methods are \textit{k} heatmap channels (where \textit{k} is the number of the landmark points), the output of CBR methods are the predicted landmark points, a $k-$dimensional vector. Thus, while the former requires complex post-processing, no post-processing is needed for the latter method. Moreover, CNNs that can be used as a backbone in CBR methods are mostly more efficient than CNNs that are used for the HBR methods since the models require greater number of model parameters to be capable of creating \textit{k} heatmap channel accurately.

\begin{figure}[t!]
 \centering
 \includegraphics[width=\columnwidth]{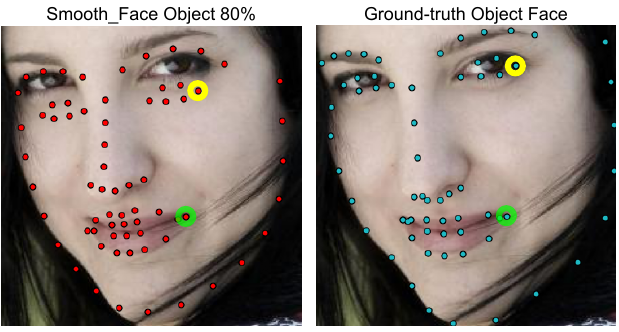}
 \caption{ We use the difference between the coordinate of a landmark point in \textit{Smooth\_Face} and its corresponding Ground\_truth point to measure how hard is the prediction of that point. Accordingly, the prediction of the \textit{Yellow} facial landmark point is harder for the model compared to the prediction of the \textit{Green} one.}
  \label{fig:main_idea}
\end{figure}

Loss functions are an integral part of each deep learning model that can affect the performance of the network dramatically. We proposed a piece-wise loss function called ACR Loss designed to find the most challenging points during the training stage and guide the network to pay more attention to them. Inspired by the Active Shape Model (ASM)~\cite{ordas2003active}, we model a face object (the ground truth landmark points for a face) using the point-wise mean of all facial landmark points plus the shape-specific variation which is generated using the Eigenvectors of Covariance matrix (see Sec.\ref{sec:Proposed_Arch}). Using more Eigenvectors will decrease the similarity between the mean face object and the ground truth face objects.
We use this characteristic and for each face object, we consider the points with fewer similarities to the mean face as the \textit{most-challenging}, and the points with the highest similarities as the \textit{least-challenging} points (see Fig.\ref{fig:main_idea}).

More clearly, for two different facial landmark points $p_1$ and $p_2$, if the distance between the location of $p_1$ and its corresponding point in Mean\_Face object is smaller than the distance between $p_2$ and its corresponding point in Mean\_Face object, prediction of the location of $p_2$ is more challenging than $p_1$. Considering this attribute, our proposed ACR Loss adapts its curvature to penalize the network for each landmark point in a face considering the hardness level of its prediction. 

We use our proposed ACR Loss function to train 3 widely used CNNs, MobileNetV2~\cite{sandler2018mobilenetv2}, EfficientNet-B0~\cite{tan2020efficientnet}, and EfficientNet-B3~\cite{tan2020efficientnet} the challenging 300W dataset ~\cite{sagonas2013300}, and the Caltech Occluded Faces in the Wild (COFW) dataset~\cite{burgos2013robust}. Our experimental results show that the accuracy of face alignment using our novel ACR Loss significantly improves the performance of the CBR models compared to the widely used L2 loss. The contributions of this paper are:
\begin{itemize}
\item We propose a method to define the hardness level of each landmark point in a face object.
\item We propose ACR Loss, which can adapt its curvature concerning the hardness level of each landmark point in a face. Accordingly, the magnitude and the influence of the loss are defined adaptively according to the hardness level of the landmark points.
\end{itemize}

The remainder of this paper is organized as follows. Sec.~\ref{sec:lit} reviews the related work in face alignment. Sec.~\ref{sec:Proposed_Arch} describes the proposed ACR Loss and how it improves the accuracy of face alignment task. Sec.~\ref{sec:experiment} provides the experimental results, and finally, Sec.~\ref{sec:conclusion} concludes the paper with some discussions on the proposed method and future research directions.  
%

\section{Related Work}\label{sec:lit}

Automated facial landmark points detection has been studied extensively by the computer vision community. Template-based fitting methods such as ASM~\cite{ordas2003active} and AAM~\cite{cootes2004statistical} are among the \textit{classical methods}, where they aim to constrain the search space by using prior knowledge. Regression-based methods consider a facial image as a vector and use a transformation such as Principle Component Analysis (PCA), Discrete Cosine Transform (DCT)~\cite{salah2007robust}, or Gabor Wavelet~\cite{arca2006face, vukadinovic2005fully} to transform the image into another domain, then a classification algorithm such as Support Vector Machines (SVM)~\cite{antonini2003independent, du2008svm} or boosted cascade detector~\cite{viola2001robust} is used to detect facial landmarks. 


\textit{CBR methods:} Sun~\etal~\cite{sun2013deep} proposed a CNN using the cascaded deep convolutional network for face alignment. A coarse-to-fine auto-encoder networks (CFAN) proposed by Zhang~\etal~\cite{zhang2014coarse} for real-time face alignment. A cascaded coordinate regression method called the mnemonic descent method (MDM) proposed by  Trigeorgis~\etal~\cite{trigeorgis2016mnemonic} for facial landmark point detection. Xiao~\etal~\cite{xiao2016robust} used a recurrent neural network while fusing the feature extraction and the regression steps and trained the network end-to-end. Two-Stage Re-initialization Deep Regression Model (TSR)~\cite{lv2017deep} splits a face into several parts to ease the parts variations snd regresses the coordinates of different parts. Valle~\etal~\cite{valle2018deeply} proposed a simple CNN for better initialization to Ensemble of Regression Trees (ERT) regressor used for face alignment. Wingloss, introduced by Feng~\etal~\cite{feng2018wing}, is a new loss function capable of overcoming the widely used L2-norm loss in conjunction with a pose-based data balancing (PDB). Zhang~\etal~\cite{zhang2019facial} proposed a weakly-supervised framework to detect facial components and landmarks simultaneously. Ning~\etal~\cite{ning2020cpu} proposed a new loss function to enhance the convergence of local regions. An attention distillation module proposed by Sadiq~\etal~\cite{sadiq2019facial} to infer the occlusion probability of each position in high-level features. Zhang~\etal~\cite{zhang2020phased} proposed a group-wise face alignment dividing the landmark points into two stages, affine transformations and non-rigid distortions. Fard~\etal~\cite{fard2021asmnet} proposed a lightweight multi-task network for jointly detecting facial landmark points as well as the estimation face pose. KD-Loss~\cite{fard2021facial} proposed a knowledge distillation-based architecture for face alignment.

\textit{HBR methods:}  Yang~\cite{yang2017stacked} proposed a two-part network including a supervised transformation to normalize faces, as well as a stacked hourglass network~\cite{newell2016stacked}, which is designed to predict heatmaps. HRNet \cite{sun2019high}, a high-resolution network applicable in many Computer Vision tasks such as facial landmark detection, achieves a reasonable accuracy.  Liu~\etal~\cite{liu2019attention} proposed an attention-guided coarse-to-fine network, which is guided to emphasize key information while suppressing less important information. In addition, Iranmanesh~\etal~\cite{iranmanesh2020robust} proposed an approach to provide a robust face alignment algorithm that copes with shape variations by aggregating a set of manipulated images to capture a robust landmark representation. Park~\etal~\cite{park2020complementary} proposed a complementary regression network to combine both global and local regression methods for face alignment. An encoder-decoder CNN~\etal~\cite{valle2020cascade} uses a cascade of CNN regressors to increase the accuracy of the face alignment task. Wu~\etal~\cite{wu2021design} used a lightweight U-Net model and a dynamic optical flow to obtain the optical flow vector of the landmark and uses dynamic routing to improve landmark stabilization. To improve the tolerance of the face alignment toward the occlusion, Park~\etal~\cite{park2021acn} combined a coordinate regression network and a heatmap regression network with spatial attention. 

Although heatmap-based models achieve better results than coordinate-based models, they perform poorly in terms of inference time, computational complexity, and memory consumption. On the contrary, the coordinate-based models do not require any post-processing operations. Hence, since the efficiency of the model is a key factor in this paper, we train 3 different models one using L2 and one using our proposed ACR Loss. We show in Sec.\ref{sec:experiment} that using ACR Loss improves the performance of face alignment compared to the widely used L2 loss.


\section{Proposed ACR Loss}\label{sec:Proposed_Arch}
In this section, we first discuss the issues with the previously proposed loss functions which are used in order to train the CBR methods, l2, and L1 losses. We then present our proposed ACR Loss and show its effectiveness in improving the performance of CBR face alignment.

\subsection{Issues with the Current CBR Loss Functions}
The loss function and the magnitude of its gradient (aka the \textit{influence}) of the loss function, are the two attributes that we analyze in our investigation. 

For l2 loss ($y = x^2$), the magnitude of the gradient is $|x|$. Accordingly, The influence of the loss function depends on the magnitude of the error. Therefore, the influence of the loss function on the network is large for large errors while it is small for small errors. The optimization problem for a single landmark point is solvable using Stochastic Gradient Descent (SGD) or numerical methods. However, when it comes to the optimization of $M \times b$ facial landmark points --\textit{M} is the number of facial landmark points defined for each input image, and \textit{b} is the size of the mini-batch--, the accumulative influence of the large errors on the loss function can be more significant than the influence of the small errors. Hence, the loss function might only focus on localizing the landmark points with large errors while ignoring the points with small errors. This characteristic of l2 loss can negatively affect the performance of facial landmark localization and result in an inaccurate prediction. 


For L1 loss ($y = |x|$), the magnitude of the gradient is $1$, indicating that the influence of the loss is agnostic to the magnitude of the error. Although L1 loss does not have the above-mentioned issue of l2 loss, its indifference on the magnitude of the errors affects the optimization problem negatively, meaning that the training of the model requires more steps to converge to the optimal solution.

To tackle the weaknesses of l2, and L1 loss functions, we propose a piece-wise loss function which is logarithmic ($y = ln(|x|^{2-\Phi} + 1)$) for small errors and quadratic for large errors. Moreover, we define $\Phi \in[0,1]$ such that it is larger for challenging points compared to points which are easier for the network to be predicted (see Sec.\ref{sec:ACR_loss}). Accordingly, $\Phi$ modifies the curvature of the loss function to adapt the loss magnitude and the influence of the loss on the model for each facial landmark point according to its hardness of prediction for the model. Thus, for the challenging points where $\Phi \approx 1$, the logarithmic piece of the ACR Loss is $y \approx ln(|x| + 1)$, while for the less challenging ones ($\Phi \approx 0$) it is $y = ln(x^2 + 1)$ (see Fig~\ref{fig:ACR_loss}). Typically, for each landmark point that exists in a face, we define the corresponding weight, $\Phi$, which is used to adapt the influence of the logarithmic part. Hence, instead of only considering the magnitude of the errors (the distance between the ground truth and the predicted landmark point) to design the loss function, we define a metric to measure how challenging a landmark point is and adapt the loss curvature accordingly.

The gradient of the logarithmic part for $\Phi \approx 1$ is $y' = \frac{1}{1+|x|}$ which means the influence of the loss increases as the errors become smaller. Consequently, contrary to l2 loss (where the influence of the loss reduces as the magnitude of the error decreases), the ACR Loss is designed to guide the network to pay more attention to the localization of the \textit{challenging} points no matter how small the magnitude of the error is. On the contrary, for less challenging landmark points, where $\Phi \approx 0$, the gradient of the logarithmic part is $y' = \frac{2 |x|}{1+x^2}$. Therefore, as the error decreases the influence of the loss decreases too. We show in Sec.\ref{sec:experiment} that this characteristic of the ACR Loss enables it to perform more accurately in face alignment tasks compared to the widely used L2 loss.

\subsection{ACR Loss} \label{sec:ACR_loss}
Inspired by ASM~\cite{cootes2004statistical}, we define a  $\mathbf{Face}$ object using an \textit{M-}dimensional vector in the following Eq.~\ref{eq:esm_deformable}:
\begin{equation} \label{eq:esm_deformable}
\mathbf{Face_{M\times1}} \approx \mathbf{Mean\_Face}_{M\times 1} + \mathbf{V}_{M\times k}~ \mathbf{b}_{k\times 1} 
\end{equation}
where \textit{M} is the number of facial landmark points defined for each image, $\mathbf{Mean\_Face}$ object is the mean shape object created by calculating the point-wise mean of all facial landmark points in the training set. In addition, $\mathbf{V}= \{v_{1}, v_{2}, ... , v_{k}\}$ contains $k$ Eigenvectors of the Covariance matrix created from all samples in the training set. In addition, $\mathbf{b}$ is a $k$ dimensional vector given by Eq.~\ref{eq:b_vector}:
\begin{equation} \label{eq:b_vector}
\mathbf{b}_{k\times 1}  = \mathbf{V}^\intercal_{ k\times M} (\mathbf{Face}_{M\times1} - \mathbf{Mean\_Face}_{M\times1})
\end{equation}
Considering that the statistical variance (\textit{i.e.}, Eigenvalue) of the $i^{th}$ parameter of the vector $\mathbf{b}$ is $\lambda_i$. To make sure the generated shape object after applying ASM (\textit{i.e.}, approximating a face object using Eq.~\ref{eq:esm_deformable}) is relatively similar to their corresponding ground truth, the parameter $b_i$ of vector $\mathbf{b}$ is usually limited to $\pm3\sqrt{\lambda_i}$~\cite{cootes2000introduction}. Then, we define the $\mathbf{Smooth\_Face}$ object as follows in Eq.~\ref{eq:smooth_sace}:
\begin{align}\label{eq:smooth_sace}
\begin{split}
{\mathbf{Smooth\_Face}}_{M\times 1}  = {\mathbf{Mean\_Face}}_{M \times 1}  + 
\mathbf{{\tilde V}}_{M\times k} \mathbf{b}_{k\times 1}
\end{split}
\end{align}
\begin{figure*}[t!]
 \centering
 \includegraphics[width=18cm]{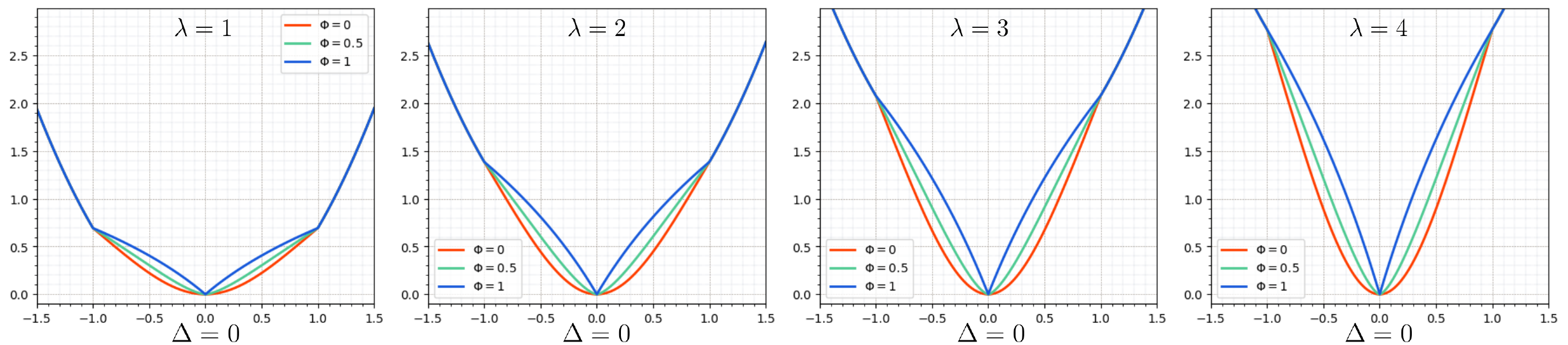}
 \caption{The proposed ACR Loss for a single landmark point. In each figure, we depicts the ACR Loss curvature for $\Phi$ as equal to 0, 1, 2. Besides, we depict the ACR Loss curvature for different values of the hyper-parameter $\lambda$ as well.}
  \label{fig:ACR_loss}
\end{figure*}

We define $l \in \{0, 1, ..., k\}$ indicates the number of the Eigenvectors we use to create the $\mathbf{Smooth\_Face}$ object. 

The number of the Eigenvectors that we choose for creating $\mathbf{Smooth\_Face}$ object defines the similarity between the original face object (the ground truth face) and the generated  $\mathbf{Mean\_Face}$ object. In other words, if we define the number of the Eigenvectors (or the parameter $l$ in Eq.~\ref{eq:esm_deformable}) equal to \textit{0},  $\mathbf{Smooth\_Face}$ object will be the same as $\mathbf{Mean\_Face}$ object. Likewise, if we use all the Eigenvectors, the generated  $\mathbf{Smooth\_Face}$ object will be the same as the original  $\mathbf{Face}$, and thus having the least similarity to  $\mathbf{Mean\_Face}$ object.

Since the variation of $\mathbf{Smooth\_Face}$ object is less than the ground truth face object, --as a rule of thumb-- it is easier for the network to learn the distribution of  $\mathbf{Smooth\_Face}$ object compared to the distribution of  $\mathbf{Face}$. We use this characteristic to propose our ACR Loss. The main idea of the ACR Loss is to find the most challenging landmark points for each input image and penalize the model more by changing the loss curvature adaptively for such points compared to less challenging points. For each input image, we use the distance between each landmark point and the corresponding points in  $\mathbf{Smooth\_Face}$ object as a metric that indicates how challenging a landmark point is. For an input image $img_i$, the ground truth face object $Face_i$, and the corresponding smooth face $Face_i$, we define $\Phi_{i,m}$ as follows in Eq.~\ref{eq:phi}:
\begin{align}\label{eq:phi}
\begin{split}
\Phi_{i,m} =  \frac{ |\mathbf{Smooth\_Face}_{i,m} - \mathbf{Face}_{i,m}|}
            { max(|\mathbf{Smooth\_Face}_{i,q} - \mathbf{Face}_{i,q}|) ~ \forall q \in M  }
\end{split}
\end{align}
such that $\mathbf{Face}_{i,m}$ is the $m^{th}$ landmark points of the $i^{th}$ sample in the training set. We use $\Phi_{i,m}$ as a \textit{weight} indicating how challenging is a landmark point, and accordingly adapt the loss curvature to declare the magnitude and the influence of the loss. Then, we define our ACR Loss in Equations~\ref{eq:delta},~\ref{eq:ACR_loss_item},~\ref{eq:ACR_loss}:
\begin{align}\label{eq:delta}
\begin{split}
    \Delta_{i,m} = |\mathbf{Face}_{i,m} - \mathbf{Pr\_Face}_{i,m}| \\
\end{split}
\end{align}
\begin{align}\label{eq:ACR_loss_item}
\begin{split}
 loss\_face_{i,m} = \left\{
        \begin{matrix} 
        \lambda \ln( 1 + \Delta_{i,m}^{2-\Phi_{i,m}} )   && If:~\Delta_{i,m}  \leq 1 \\ \\
        \Delta_{i,m}^2 + C     && If:~\Delta_{i,m}  > 1 
        \end{matrix}\right.
\end{split}
\end{align}
\begin{align}\label{eq:ACR_loss}
\begin{split}
Loss_{ACR} = \frac{1}{M~N}\sum_{i=1}^{N} \sum_{m=1}^{M} loss\_face_{i,m}
\end{split}
\end{align}
where, \textit{M} and \textit{N} are the numbers of the landmark points and images in the training set, respectively. $Pr\_Face_{i,m}$ is the $m^{th}$ landmark point of the $i^{th}$ the predicted face object, and $C = \Phi_{i,m} \ln(2) - 1$ is a constant defined to link the quadratic part to the logarithmic part smoothly. In addition, the hyper-parameter $\lambda$ is defined to adjust the ACR Loss curvature (see the ablation study, Sec~\ref{sec:ablation_study}, for the effect of the different value of $\lambda$ on the performance of the face alignment task.)

As Fig.~\ref{fig:ACR_loss} depicts, we define the ACR Loss as a piece-wise function to make it cope well with both small and large errors. For the large errors (where $\Delta_{i,m}$ is greater than 1), we define the ACR Loss to be a quadratic function. Accordingly, the magnitude and the influence of the loss function rely upon the magnitude of the error. For small errors (where $\Delta_{i,m}$ is lower than or equal to 1), we modify the loss curvature according to the value of $\Phi_{i,m}$, which indicate the \textit{hardness} level of the prediction of the corresponding point. Accordingly, for the challenging points, the value of $\Phi_{i,m}$ is close to zero, the magnitude of the loss and its influence decrease as the value of the error decreases. In contrast, for the challenging points, the value of $\Phi_{i,m}$ is close to $1$ and accordingly, the ACR Loss becomes a logarithmic loss function. Consequently, the influence of the loss function increases as the value of the error decreases. 

\begin{figure}[t!]
  \centering
  \includegraphics[width=\columnwidth]{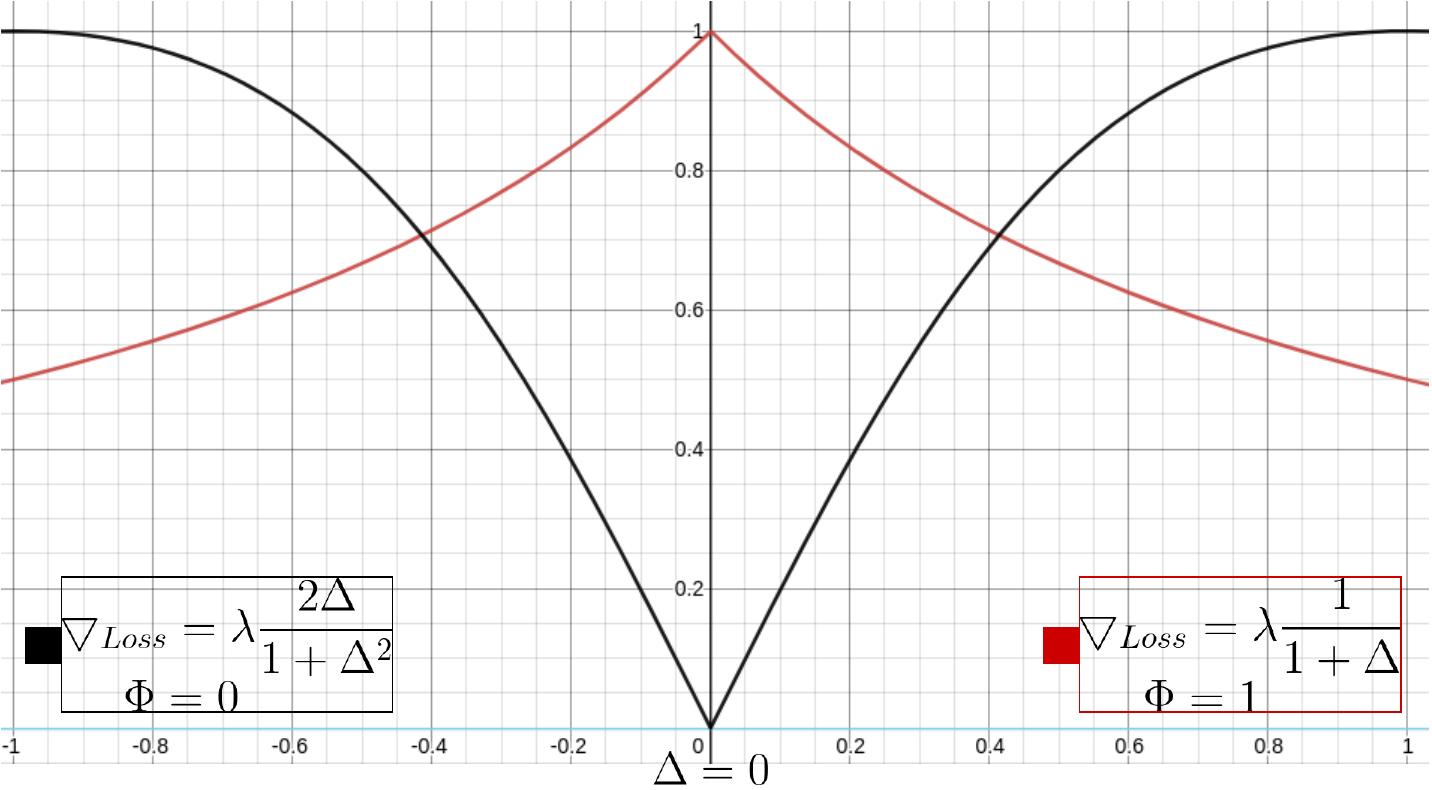}
  \caption{Gradient of the ACR Loss function for the small errors where $\Delta \leq 1 $. For  $\Phi=1$, the influence of the loss decreases as the error decreases, while for $\Phi=0$, the influence of the loss increases as the error magnitude decreases.}
  \label{fig:gradient_function}
\end{figure}
\newcommand{\STAB}[1]{\begin{tabular}{@{}c@{}}#1\end{tabular}}

\begin{table*}[t]
\caption{Comparing NME (in \%), FR (in \%), and AUC of landmarks localization using ACR loss and L2 loss on 300W~\cite{sagonas2013300}, and COFW~\cite{burgos2013robust} datasets.}
 \label{tbl:model_result}
\centering
\small
\resizebox{17cm}{!}{
\begin{tabular}{c|cccc|cccc|cccc}
\hline 
\multirow{3}{*}{Method} & \multicolumn{4}{c|}{NME~($\downarrow$)}                                                                             &\multicolumn{4}{c|}{FR~($\downarrow$)}
                        & \multicolumn{4}{c}{AUC~($\uparrow$)}
                        \\ \cline{2-13} 
                        & \multicolumn{1}{c|}{\multirow{2}{*}{COFW}} & \multicolumn{3}{c|}{300W}   & \multicolumn{1}{c|}{\multirow{2}{*}{COFW}} & \multicolumn{3}{c|}{300W}   & \multicolumn{1}{c|}{\multirow{2}{*}{COFW}} & \multicolumn{3}{c}{300W}    \\ \cline{3-5} \cline{7-9} \cline{11-13} 
                        & \multicolumn{1}{c|}{}                      & Challenging & Common & Full & \multicolumn{1}{c|}{}                      & Challenging & Common & Full & \multicolumn{1}{c|}{}                      & Challenging & Common & Full \\ \hline 
\hline

MN\textsubscript{base}  
                    & 4.93      &7.32       & 4.22      & 4.82     
                    & 0.59      &12.59      & 0.18      & 2.61                      
                    & 0.7338    &0.4953     & 0.7896    & 0.7319      \\
MN\textsubscript{ACR}
                    & 3.78      & 6.16      & 3.81      & 4.27  
                    & 0.39      & 1.48      & 0.00      & 0.29
                    & 0.8215    & 0.6076    & 0.8197    & 0.7781       \\ \hdashline

EF-0\textsubscript{base}
                    & 4.93      & 6.92      & 4.67      & 5.11 
                    & 1.18      & 7.40      & 0.18      & 1.59  
                    & 0.7333    & 0.5228    & 0.7535   & 0.7083      \\
EF-0\textsubscript{ACR}
                    & 4.20      & 6.71      & 4.38      & 4.83 
                    & 0.59      & 4.44      & 0.18      & 1.01 
                    & 0.7892    &0.5426     & 0.7785    & 0.7323      \\ \hdashline

EF-3\textsubscript{base}
                    &3.71       &6.01      &3.81       & 4.24  
                    &0.39       &1.48      &0.00       & 0.29  
                    &0.8275     &0.6196    &0.8215     & 0.7820     \\ 
EF-3\textsubscript{ACR}
                    &3.47       &5.36       &3.36       & 3.75 
                    &0.39       &1.48       &0.00       & 0.29     
                    &0.8421     &0.6838     &0.8504    & 0.8177     \\ \hline 
\end{tabular} 
 }
\end{table*}

In addition, in L2 loss the magnitude of the loss value \textit{only} relies on the prediction error, which is the distance between a ground truth facial landmark point and its corresponding predicted point. Therefore, the greater the prediction error, the greater the loss value. Consequently, L2 loss is less sensitive to small errors compared to the large errors.

To cope with this issue, for each landmark point $P_{g}$, we define $\Phi$ as the distance between the landmark point $P_{s}$ in $\mathbf{Smooth\_Face}$ object and the corresponding landmark point $P_{g}$ in the ground truth set. $\Phi$ can modify the curvature of the ACR Loss, so the magnitude of the loss is set based on the hardness of the prediction of $P_{g}$ for the network. Consequently, the ACR Loss guides the model to pay more attention to localizing the challenging points which results in accuracy improvement.


The number of the Eigenvectors (or the parameter $k$) defines the similarity between the ground truth face object and the corresponding Smooth\_Face. In the very first epochs during the training, the predicted face objects are similar to the Mean\_Face object. Then, gradually the model learns the face-specific features so that the predicted facial landmark point becomes more similar to their corresponding ground-truth points. Accordingly, we increase the number of the Eigenvectors for creating the $\mathbf{Smooth\_Face}$ objects. In other words, as the prediction accuracy of the model increases during the training, we need to add more face-specific variations (see Eq.\ref{eq:smooth_sace}) to the generated Smooth\_Faces to make sure that $\Phi_{i,m}$ represent the most challenging points. Empirically we choose $k$ to be $k=80\%$ of all the available Eigenvectors from epoch 0 to 15, $k=85\%$ from epoch 16 to 30, $k=90\%$ from epoch 31 to 70, $k=95\%$ from epoch 71 to 100, and $k=97\%$ from epoch 101 to 150. So, in the very first epochs, we update $k$ faster comparing to the last epochs since on the last epochs the network required more effort to predict the facial landmark points more accurately.
%
\subsection{ACR Loss Optimization}
We use SGD algorithm to optimize our proposed ACR Loss. Based on the Eq.~\ref{eq:ACR_loss} (and as also Fig.~\ref{fig:ACR_loss} shows), ACR Loss is a convex function so that we can use SGD to find the optimal minimum. Therefore, we calculate the gradient of our piece-wise ACR Loss in Eq.\ref{eq:gradient}:
\begin{align}\label{eq:gradient}
\begin{split}
 \nabla Loss_{ACR} = \left\{
        \begin{matrix} 
         \lambda\frac{\Delta(\Phi-2)}{\Delta^\Phi +\Delta^2 }   && If:~\Delta  \leq 1 \\ \\
        2~\Delta    && If:~\Delta  > 1 
        \end{matrix}\right.
\end{split}
\end{align}
Accordingly, for $\Delta  > 1 $ part, the derivative of the loss function is $2~\Delta$ which is differentiable on its domain of definition, and SGD can find its optimal minimum. As Fig.\ref{fig:gradient_function} depicts, for $\Delta \leq 1 $ part, the gradient function is $\lambda\frac{\Delta(\Phi-2)}{\Delta^\Phi +\Delta^2 }$. Therefore, for $\Phi=0$, the gradient of the loss is $\lambda\frac{2\Delta}{1 +\Delta^2 }$, a differentiable function in its domain of definition. Likewise, for $\Phi=1$, the gradient of the loss is $\lambda\frac{1}{1 +\Delta }$, which is differentiable in its domain of definition as well. Thus, SGD is capable of finding its
optimal minimum.
 
\section{Experimental Results} \label{sec:experiment}
For the ease of use, we define MN\textsubscript{ACR}, EF-0\textsubscript{ACR}, and EF-3\textsubscript{ACR} which are MobileNetV2~\cite{sandler2018mobilenetv2}, EfficientNet-B0~\cite{tan2020efficientnet}, and EfficientNet-B3~\cite{tan2020efficientnet} being trained using our ACR Loss versus the corresponding MN\textsubscript{base},EF-0\textsubscript{base}, and EF-3\textsubscript{base} being trained using the widely used L2 loss. We uses 300W~\cite{sagonas2013300}, and COFW~\cite{burgos2013robust} dataset for our experiments.

\section{Implementation Details}
For the training set in each dataset, we cropped all the images and extract the face region. Then the facial images are scaled to $224\times224$ pixels. We augment the images (in terms of contrast, brightness, and color) to add robustness of data variation to the network. We train networks for about 150 epochs, using the Adam optimizer~\cite{kingma2014adam} with a learning rate of $10^{-3}$, $\beta_1 = 0.9$, $\beta_2 = 0.999$, and $decay = 10^{-5}$ on a NVidia 1080Ti GPU.

%
\begin{figure}[t]
  \centering
  \includegraphics[width=\columnwidth]{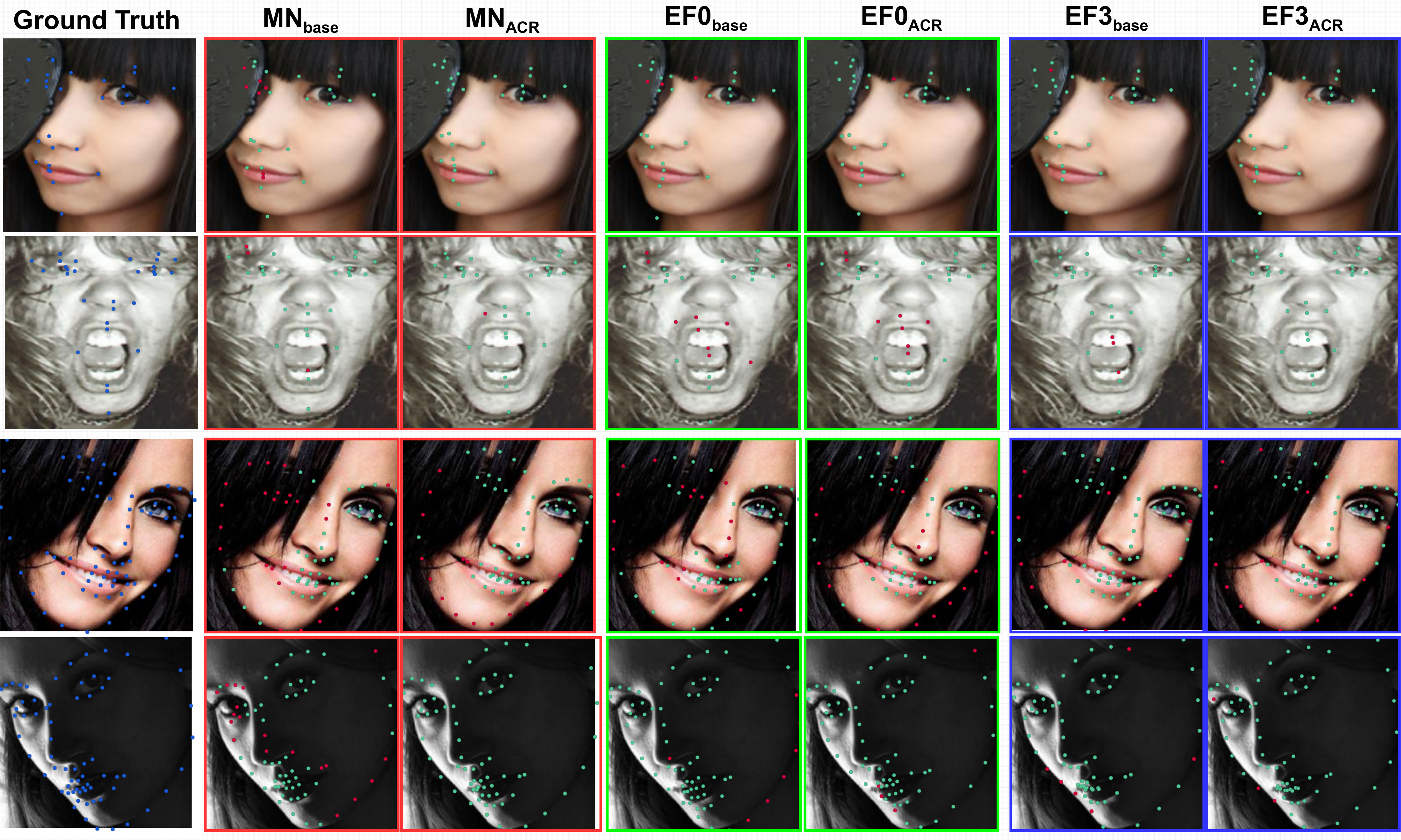}
  \caption{Examples of facial landmark points detection using our proposed ACR loss versus l2 loss on COFW~\cite{burgos2013robust}(first and second rows) and 300W~\cite{sagonas2013300} (third and forth rows) dataset. For each landmark point, if the error rate is more than $0.1$, it is considered as a failure and we printed it as red, and otherwise it is green.}.
  \label{fig:ACR_samples}
\end{figure}
\begin{table}[t!]
\caption{NME (in \%) and FR (in \%) of 28-point landmarks localization on COFW~\cite{burgos2013robust} dataset.}
\label{tbl:tbl_results_cofw}
\centering
{
\begin{tabular}{l c c}
\hline
Method                                      & NME                   & FR  \\ \hline
SFPD~\cite{wu2017simultaneous}              & 6.40                  & -             \\ 
DAC-CSR~\cite{feng2017dynamic}              & 6.03                  & 4.73          \\ 
FARN~\cite{hou2018face}                     & 5.81                 & -             \\ 
CNN6 (Wing + PDB)~\cite{feng2018wing}       & 5.44                  & 3.75          \\ 
ResNet50 (Wing + PDB)~\cite{feng2018wing}   & 5.07                  & 3.16          \\ 
LAB~\cite{wu2018look}                       & 3.92                  & 0.39          \\ 
ODN~\cite{zhu2019robust}                    & 5.30                  & -             \\ 
ResNet50-FFLD~\cite{yan2020fine}            & 5.32                  & -             \\ 
ACN~\cite{park2021acn}                      & 3.83                  & -             \\ 

\multicolumn{3}{l}{}
\\[-1em] \hline

EF-3\textsubscript{ACR} (\textbf{ours})        & \textbf{3.47}                    & \textbf{0.39}    \\ 

\hline
\end{tabular}
}
\end{table}

%

\subsection{Evaluation on COFW}
As shown in Table~\ref{tbl:model_result}, for the COFW~\cite{burgos2013robust} dataset, using MobileNetV2~\cite{sandler2018mobilenetv2} as the network, the NME and FR reduce $1.15\%$ (from $4.93$ to $3.78$) and $0.2\%$ (from $0.59\%$ to $0.39\%$) respectively, and ACU increases about $8.77\%$ from $0.7338$ to $0.8215$ while we train the model using ACR Loss. Likewise, using EfficientNet-B0~\cite{tan2020efficientnet} as the backbone network, we see that both NME and FR fell from $4.93\%$ to $4.20\%$ ($0.73\%$ reduction), and from $1.18\%$ to $0.59\%$ ($0.59\%$ reduction) respectively, and ACU rises about $5.59\%$ from $0.7333$ to $0.7892$. As well as that, using EfficientNet-B3~\cite{tan2020efficientnet}, we also face about $0.24\%$ (from $3.71\%$ to $3.47\%$) reduction on NME, while AUC increases from $0.8275$ to $0.8421$, about $1.46\%$.

Table~\ref{tbl:tbl_results_cofw} shows the results of the state-of-the-art methods as well as $EF3\_{ACR}$, the EfficientNet-B3~\cite{tan2020efficientnet} trained using our proposed ACR Loss. As shown in Table~\ref{tbl:tbl_results_cofw}, the NME of $EF3\_{ACR}$ is $3.47\%$, which is the lowest among the state-of-the-art methods. In addition, the FR of our method is $0.39\%$, which equals to the FR of LAB~\cite{wu2018look} and the lowest as well. Fig~\ref{fig:ACR_samples} shows some samples of face alignment using ACR Loss compared to l2 loss.

%
\subsection{Evaluation on 300W}

Table~\ref{tbl:model_result} shows that on the \textit{Challenging} subset of 300W~\cite{sagonas2013300} dataset, training the networks using ACR Loss results in about $1.16\%$ (from $7.32\%$ to $6.16\%$), $0.21\%$ (from $6.92\%$ to $6.71\%$) and $0.65\%$ (from $6.01\%$ to $5.36\%$) reduction in NME using MobileNetV2~\cite{sandler2018mobilenetv2}, EfficientNet-B0~\cite{tan2020efficientnet}, and EfficientNet-B3~\cite{tan2020efficientnet} as the network respectively. Similarly, the FR decreases about $11.11\%$ (from $12.59\%$ to $1.48\%$) for MobileNetV2~\cite{sandler2018mobilenetv2}, about $2.96\%$ (from $7.40\%$ to $4.44\%$) EfficientNet-B0~\cite{tan2020efficientnet}, while it remains without any change for EfficientNet-B3~\cite{tan2020efficientnet}. Moreover, The AUC increase about $11.23\%$ (from $0.4953$ to $0.6076$), $1.98\%$ (from $0.5228$ to $0.5426$) and $6.42\%$ (from $0.6196$ to $0.6838$), using MobileNetV2~\cite{sandler2018mobilenetv2}, EfficientNet-B0~\cite{tan2020efficientnet}, and EfficientNet-B3~\cite{tan2020efficientnet} as the network respectively and train the models using ACR Loss compared to L2 loss.

As Table~\ref{tbl:tbl_results_300w} shows, the performance of the EF-3\textsubscript{ACR} is comparable to the state-of-the-art methods specifically on the \textit{Challenging} subset of the 300W~\cite{sagonas2013300} dataset. To be more specific, on the \textit{Challenging} subset, while the lowest NME, $5.15\%$, achieved by CHR2c~\cite{valle2020cascade}, EF-3\textsubscript{ACR} achieves $5.36\%$, which is much better than many recently-proposed methods for face alignment. Accordingly, we can conclude that ACR Loss performs much better in localizing the faces under challenging circumstances such as occlusions, extreme pose, and illumination. Moreover, Table~\ref{tbl:tbl_results_300w_small} presents the NME of the lightweight state-of-the-art models with MN\textsubscript{ACR}, which is MobileNetV2~\cite{sandler2018mobilenetv2} trained using ACR Loss. According to the Table~\ref{tbl:tbl_results_300w_small}, MN\textsubscript{ACR} achieves by far the best performance. Fig~\ref{fig:ACR_samples} shows samples of face alignment using ACR Loss comparing to l2 loss.

%
\begin{table}[t!] 
\caption{NME (in \%) of 68-point landmarks localization on 300W~\cite{sagonas2013300} dataset.}
\label{tbl:tbl_results_300w}
\centering
\small

\resizebox{6.5cm}{!}{

{\begin{tabular}{l c c c }
\hline 
\multirow{2}{*}{Method} & \multicolumn{3}{c}{NME}       \\ \cline{2-4} 
                                 & Common & Challenging  & Fullset \\ \hline
FARN~\cite{hou2018face}                 & 4.23              & 7.53        & 4.88    \\ 
SAN~\cite{dong2018style}                & 3.34              & 6.60        & 3.98    \\ 
LAB~\cite{wu2018look}                   & 2.98              & 5.19        & 3.49    \\ 
ODN~\cite{zhu2019robust}                & 3.56              & 6.67        & 4.17    \\ 
HORNet~\cite{zhen2020heterogenous}      & 3.38              & 6.36         & 3.96    \\
CHR2c ~\cite{valle2020cascade}          & 2.85              & \textbf{5.15}        & 3.30    \\
mnv2\textsubscript{KD}~\cite{fard2021facial}        & 3.56              & 6.13              & 4.06       \\ 
\hline  

EF-3\textsubscript{ACR} (\textbf{ours})    & 3.36             & 5.36         & 3.75

\\ \hline
\end{tabular}}}
\end{table}
\subsection{Ablation Study}\label{sec:ablation_study}
As Fig\ref{fig:ACR_loss} depicts, the parameter $\lambda$ in Eq~\ref{eq:ACR_loss_item} can adjust the curvature of the ACR Loss. Typically, by increasing the value of $\lambda$, the magnitude of the ACR Loss function increases as well. We conducted 6 experiments to study the effect of the different values of the hyper-parameter $\lambda$ on the accuracy of the face alignment. In our experiments, we use EfficientNet-B3~\cite{tan2020efficientnet} as the model, and conduct the experiments on 300W~\cite{sagonas2013300} dataset. 

For the \textit{Common} and \textit{Full} subsets, defining $\lambda$ to be in \{ 1, 2, 3\} does not affect the NME of the face alignment too much. To be more detailed, for $\lambda=1$, the NME on the \textit{Common} subset is $3.59\%$, while it reduces to $3.51\%$ for $\lambda=2$, and then increases to $3.56\%$ for $\lambda=3$. In contrast, for the \textit{Challenging} subset, increasing the value of $\lambda$ from $1$ to $4$ results in reduction in the NME from $3.56\%$ (for $\lambda=1$) to $5.36\%$ (for $\lambda=4$). Similarly, on the \textit{Common} and \textit{Full} subsets the minimum value of the NME achieved by defining $\lambda=4$. Then, we continue increasing the value of $\lambda$ to be $5$ and then $10$. As Fig\ref{fig:ACR_loss} shows, the NME goes up on all subsets as we increase the value of $\lambda$ to be $4$. Likewise, increasing $\lambda$ to $10$ worsen the accuracy and the NME goes up to its maximum in the experiment range. Therefore, according to our experiments we define $\lambda=4$ since it results in least NME value. 

\begin{table}[t!] 
\caption{Comparison of the NME (in \%) of lightweight models in landmarks localization on 300W~\cite{sagonas2013300} dataset.}
\label{tbl:tbl_results_300w_small}
\centering
\small
\resizebox{7cm}{!}{
{\begin{tabular}{l c c c }
\hline 
\multirow{2}{*}{Method} & \multicolumn{3}{c}{NME}       \\ \cline{2-4} 
                                            & Common & Challenging  & Fullset \\ \hline
                     & \multicolumn{3}{c}{inter-ocular normalization}  \\
                                            
res\_loss~\cite{ning2020cpu}                 & -                 & -             & 4.93 \\ 
ASMNet~\cite{fard2021asmnet}                & 4.82              & 8.2           & 5.50    \\ 
MobileNet+ASMLoss~\cite{fard2021asmnet}     & 3.88              & 7.35          & 4.59    \\ 
MN\textsubscript{ACR}\textbf{(ours)}       & \textbf{3.81}  &\textbf{6.16}   & \textbf{4.27} \\

\hline 
                    & \multicolumn{3}{c}{inter-pupil normalization}  \\

DOF~\cite{wu2021design}                  & \textbf{4.86}  & 9.13   & \textbf{5.55} \\
MN\textsubscript{ACR}\textbf{(ours)}     & 5.32  &\textbf{8.94}             & 6.03
\\ \hline
\end{tabular}}}
\end{table}
\begin{figure}[t]
  \centering
  \includegraphics[width=6.5cm]{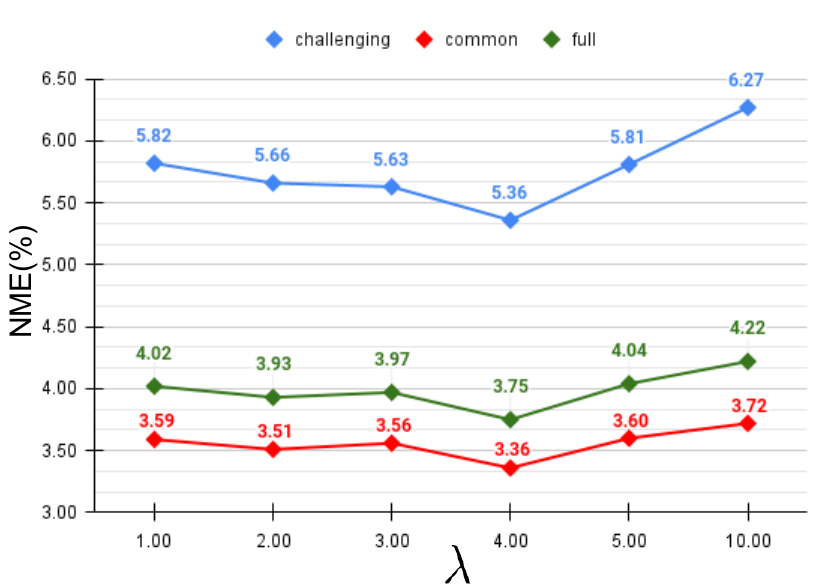}
  \caption{Studying the effect of the hyper-parameter $\lambda$ on the accuracy of the face alignment. We use EfficientNet-B3~\cite{tan2020efficientnet} trained using ACR Loss as the model and 300W~\cite{sagonas2013300} as the dataset.}.
  \label{fig:ablation_chart}
\end{figure}

\section{Conclusion and Future Work} \label{sec:conclusion}
This paper proposed an adaptive piece-wise loss function called ACR Loss. Comparing the face objects with their corresponding smoothed versions, we define a metric that indicates how challenging the prediction of a landmark point for the CNN is. Thus, for each landmark point, while l2 loss just uses the magnitude of the error to define the magnitude of the loss, ACR Loss adapts its curvature to set the loss influence and its magnitude based on the hardness of the landmark points. ACR Loss is capable of guiding the network towards focusing on the more challenging points, having the least similarity to their corresponding \textit{Smooth-Face}, compared to the less challenging points, having the highest level of similarity to their corresponding \textit{Smooth-Face}. For further study, we plan to use our proposed loss function in other computer vision tasks such as human body joint tracking.




\newpage 
\bibliographystyle{IEEEtran}
\bibliography{references}

\newpage 

\section{Supplementary Materials} \label{sec:supplementary_materials}

\section{Datasets} \label{sec:training}
300W dataset: We followed the protocol described in~\cite{ren2014face} to train our networks on the 300W~\cite{sagonas2013300} dataset. We used 3,148 faces consisting of 2,000 images from the training subset of the HELEN~\cite{le2012interactive} dataset, 811 images from the training subset of the LFPW~\cite{belhumeur2013localizing} dataset, and 337 images from the full set of the AFW~\cite{zhu2012face} dataset with a 68-point annotation. For testing, 300W~\cite{sagonas2013300} has three subsets: Common subset with 554 images, \textit{Challenging} subset with 135 images, and Full subset, including both the \textit{Common} and \textit{Challenging} subsets, with 689 images. More specifically, the \textit{Challenging} subset is the IBUG~\cite{sagonas2013300} dataset, while the Common subset is a combination of the HELEN test subset (330 images) and the LFPW test subset (224 images).

COFW dataset: This dataset contains facial images with large pose variations as well as heavy occlusions and is known as the most challenging and common issue in the real situation~\cite{burgos2013robust}. The training set contains 1345 faces, while the testing set has 507 faces. Each image in the COFW~\cite{burgos2013robust} dataset has 29 manually annotated landmarks. 


\subsection{Evaluation Metrics} \label{sec:vel_metrics}
We follow the previous work on facial landmark points detection and employ normalized mean error~(NME) to measure our model's accuracy (see Equations~\ref{eq:Mean_Error_j} and~\ref{eq:normalized_mean_error}). We define the normalizing factor, followed by MDM~\cite{trigeorgis2016mnemonic} and~\cite{sagonas2013300} as “inter-ocular” distance (the distance between the outer-eye-corners). We also calculate failure rate~(FR), defined as the proportion of failed detected faces, for a maximum error of $0.1$. Cumulative Errors Distribution~(CED) curvature and the area-under-the-curvature~(AUC)~\cite{yang2015empirical} are reported as well. 

\setlength{\tabcolsep}{0.5em}
\begin{equation} \label{eq:Mean_Error_j}
 ME_j =\frac{1}{n}\sum_{j=1}^{n} \sqrt{(P_x^i, - G_x^i)^{2} + (P_y^i - G_y^i)^{2}} 
\end{equation}
\begin{equation} \label{eq:normalized_mean_error}
 NME =\frac{1}{m \times d}  \sum_{i=1}^{m}  ME_i
\end{equation}

\section{Additional Experiments} \label{sec:experiment}
\begin{figure*}[t]
  \centering
  \includegraphics[width=15cm]{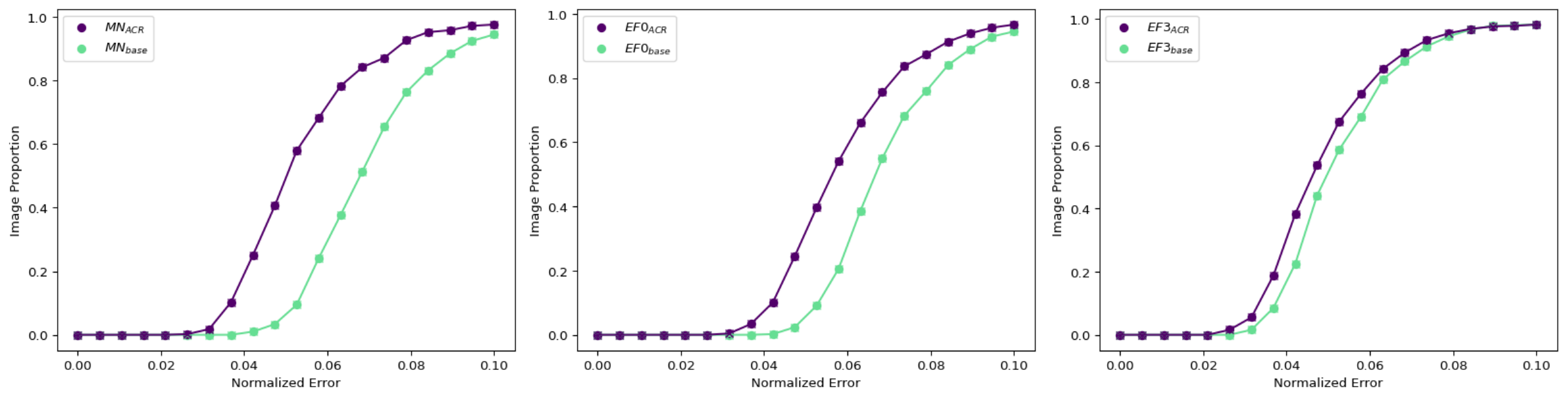}
  \caption{CED curvature for the MN\textsubscript{ACR}, EF0\textsubscript{ACR}, and EF3\textsubscript{ACR} versus MN\textsubscript{base}, EF0\textsubscript{base}, and EF3\textsubscript{base} on COFW~\cite{burgos2013robust} dataset}.
  \label{fig:ACR_CED_cofw}
\end{figure*}

\begin{figure*}[t]
  \centering
  \includegraphics[width=15cm]{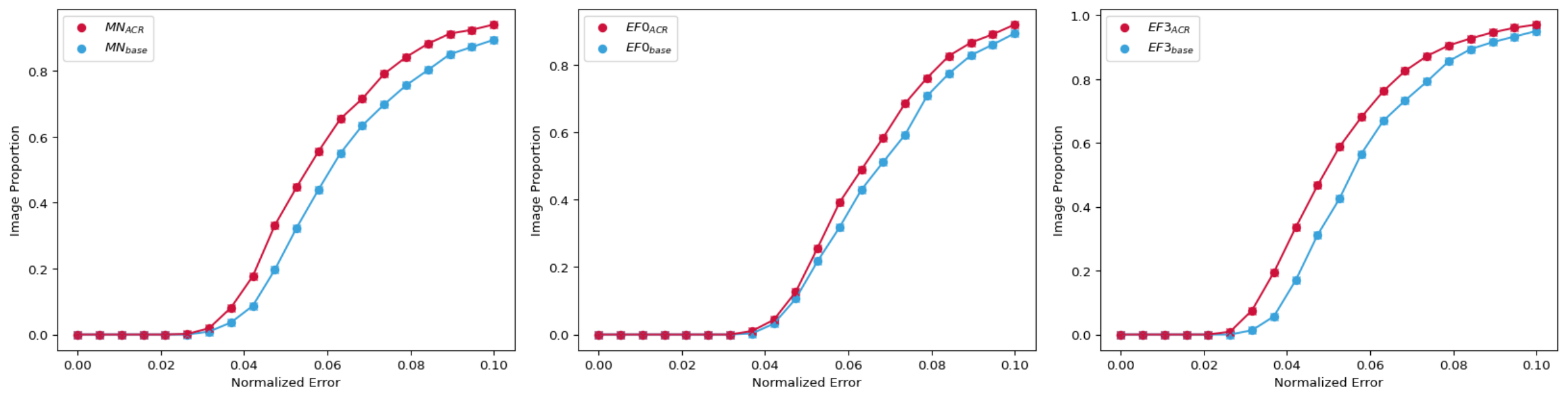}
  \caption{CED curvature for the MN\textsubscript{ACR}, EF0\textsubscript{ACR}, and EF3\textsubscript{ACR} versus MN\textsubscript{base}, EF0\textsubscript{base}, and EF3\textsubscript{base} on 300W~\cite{sagonas2013300} dataset}.
  \label{fig:ACR_CED_300w}
\end{figure*}
\begin{figure}[t]
  \centering
  \includegraphics[width=\columnwidth]{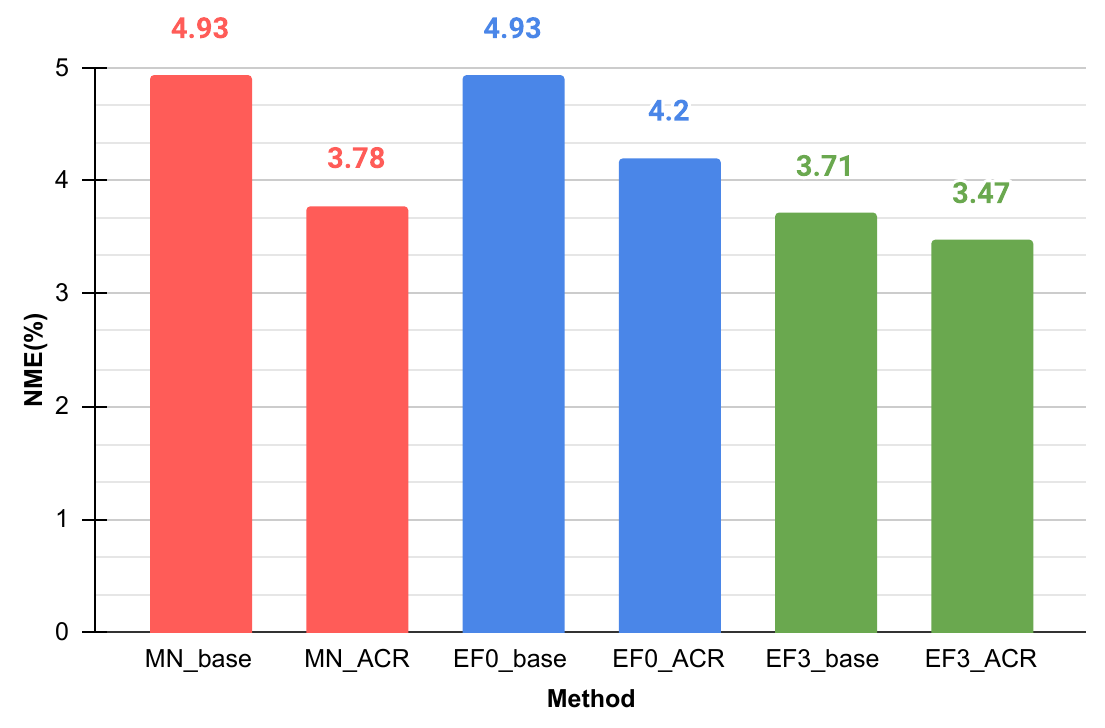}
  \caption{Comparison of NME (in \%) using MobileNetV2-\cite{sandler2018mobilenetv2}, EfficientNet-B0~\cite{tan2020efficientnet}, and EfficientNet-B3~\cite{tan2020efficientnet} as the CNN model on COFW~\cite{burgos2013robust} dataset. We train each model two times, once using L2 loss and call the models MN\_base, EF0\_base, and EF3\_base, and once using ACR loss and calling the models MN\_ACR, EF0\_ACR, and EF3\_ACR.}
  \label{fig:cofw_result_compare}
\end{figure}
\begin{figure}[t]
  \centering
  \includegraphics[width=\columnwidth]{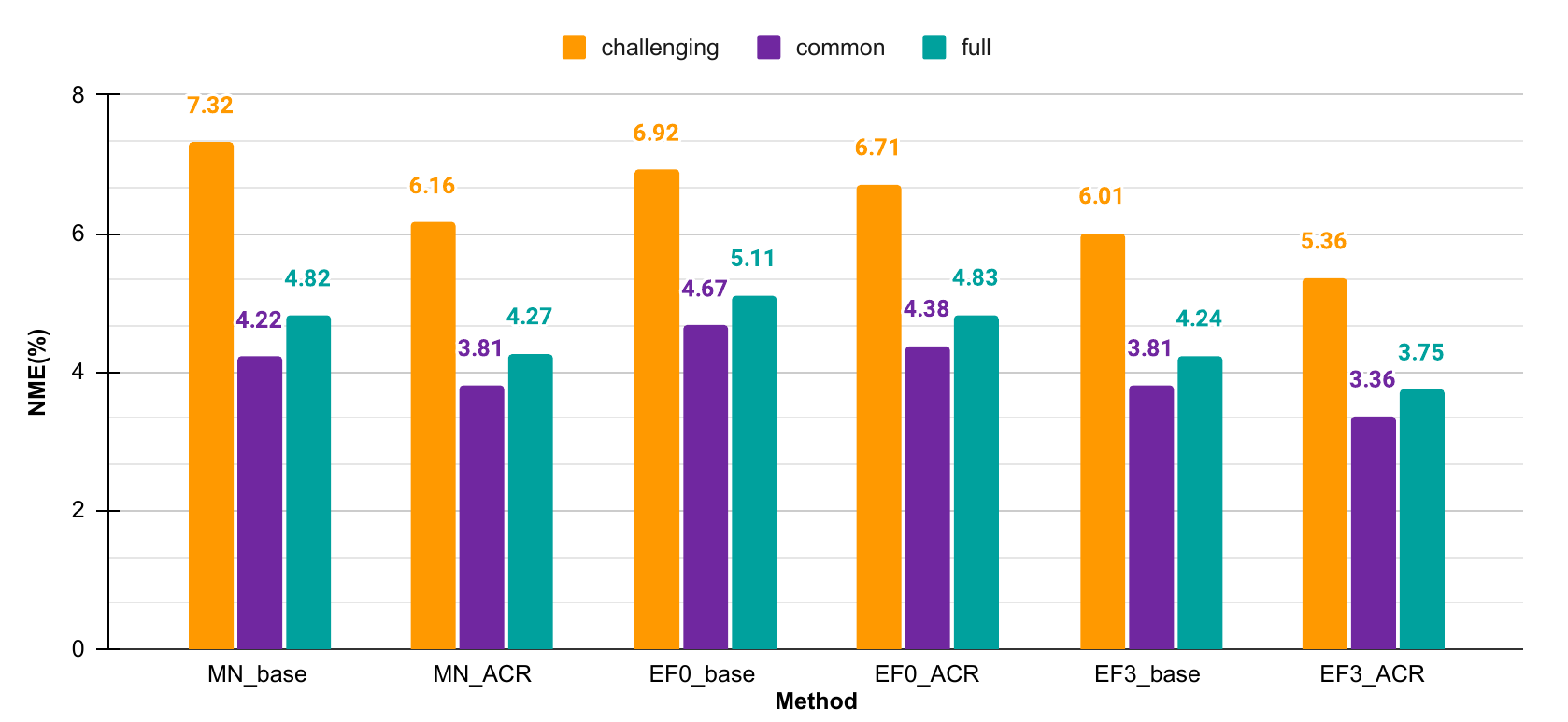}
  \caption{Comparison of NME (in \%) using MobileNetV2-\cite{sandler2018mobilenetv2}, EfficientNet-B0~\cite{tan2020efficientnet}, and EfficientNet-B3~\cite{tan2020efficientnet} as the CNN model on the 3 subsets of 300W~\cite{sagonas2013300} dataset. We train each model two times, once using L2 loss and call the models MN\_base, EF0\_base, and EF3\_base, and once using ACR loss and calling the models MN\_ACR, EF0\_ACR, and EF3\_ACR.}.
  \label{fig:300w_result_compare}
\end{figure}
%


\subsection{Evaluation on COFW}
In order to study the affect of ACR loss in different models, we provide Fig~\ref{fig:cofw_result_compare} which depicts the NME of the face alignment using MobileNetV2-\cite{sandler2018mobilenetv2}, EfficientNet-B0~\cite{tan2020efficientnet}, and EfficientNet-B3~\cite{tan2020efficientnet} COFW~\cite{burgos2013robust} dataset. According to the Fig~\ref{fig:cofw_result_compare}, on MobileNetV2-\cite{sandler2018mobilenetv2} the reduction in the value of the NME (which are $4.93\%$ using L2 loss compared to $3.78\%$ training network using ACR loss) is the greatest, about $1.15\%$. Using EfficientNet-B0~\cite{tan2020efficientnet}, the NME reduction is less, about $0.73\%$. Likewise, we experience the least reduction on EfficientNet-B0~\cite{tan2020efficientnet}, about $0.24\%$. In fact, we can conclude from Fig~\ref{fig:cofw_result_compare} that on COFW~\cite{burgos2013robust} dataset, our ACR loss is more effective for a lightweight network such as MobileNetV2-\cite{sandler2018mobilenetv2}, compared to a more heavyweight network like EfficientNet-B3~\cite{tan2020efficientnet}.

Since both NME and FR are sensitive to the outliers, we also depict the CED curvature of the MobileNetV2-\cite{sandler2018mobilenetv2}, EfficientNet-B0~\cite{tan2020efficientnet}, and EfficientNet-B3~\cite{tan2020efficientnet} trained once using L2 loss and once using ACR loss. As Fig.~\ref{fig:ACR_CED_cofw} shows, the generated CED curvature on COFW~\cite{burgos2013robust} indicates that of training the models using ACR loss (MN\textsubscript{ACR}, EF-0\textsubscript{ACR}, and EF-3\textsubscript{ACR})
perform much better than base models (MN\textsubscript{ACR}, EF-0\textsubscript{ACR}, and EF-3\textsubscript{ACR}) indicating the fact that the proposed ACR loss has effectively lead the networks to better learn the face alignment task.

%

\subsection{Evaluation on 300W}
According to Fig~\ref{fig:300w_result_compare}, which depicts the NME of the face alignment using MobileNetV2-\cite{sandler2018mobilenetv2}, EfficientNet-B0~\cite{tan2020efficientnet}, and EfficientNet-B3~\cite{tan2020efficientnet} on 300W~\cite{sagonas2013300} dataset, training the lightweight MobileNetV2-\cite{sandler2018mobilenetv2} using ACR loss results in the highest reduction in NME compared to the NME reduction taking place for EfficientNet-B0~\cite{tan2020efficientnet}, and EfficientNet-B3~\cite{tan2020efficientnet}. This characteristic of ACR loss that it can adaptively change its curvature according to the hardness of the prediction of a landmark point, can be considered more effective for the lightweight models than the heavyweight networks. 
\begin{figure*}[t!]
  \centering
  \includegraphics[width=15cm]{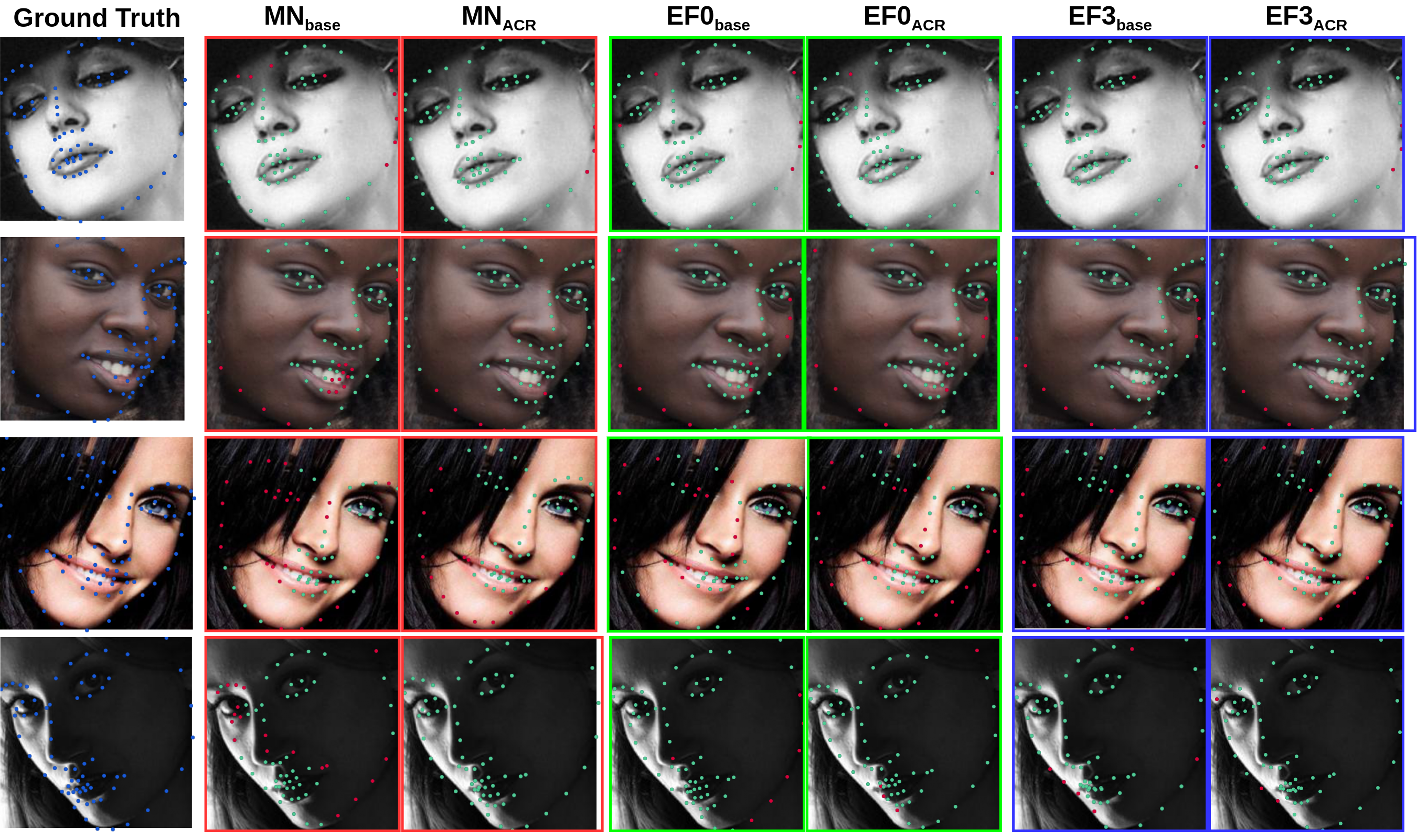}
  \caption{Examples of facial landmark points detection using our proposed ACR loss versus the L2 loss on 300W~\cite{sagonas2013300} dataset. For each landmark point, if the error rate is more than $0.1$, it is considered as a failure and we printed it as red, and otherwise it is green.}.
  \label{fig:ACR_300w_samples}
\end{figure*}

In addition, based on Fig.~\ref{fig:ACR_CED_300w} which shows CED curvature of the MobileNetV2-\cite{sandler2018mobilenetv2}, EfficientNet-B0~\cite{tan2020efficientnet}, and EfficientNet-B3~\cite{tan2020efficientnet} trained once using L2 loss and once using ACR loss, training the models using ACR loss (MN\textsubscript{ACR}, EF-0\textsubscript{ACR}, and EF-3\textsubscript{ACR}) improves the accuracy of the face alignment task compared to the base models (MN\textsubscript{ACR}, EF-0\textsubscript{ACR}, and EF-3\textsubscript{ACR}), conveying that ACR loss is capable of leading the models to focus more on localizing the challenging points and thus performs much better than the L2 loss.

\begin{figure*}[t!]
  \centering
  \includegraphics[width=15cm]{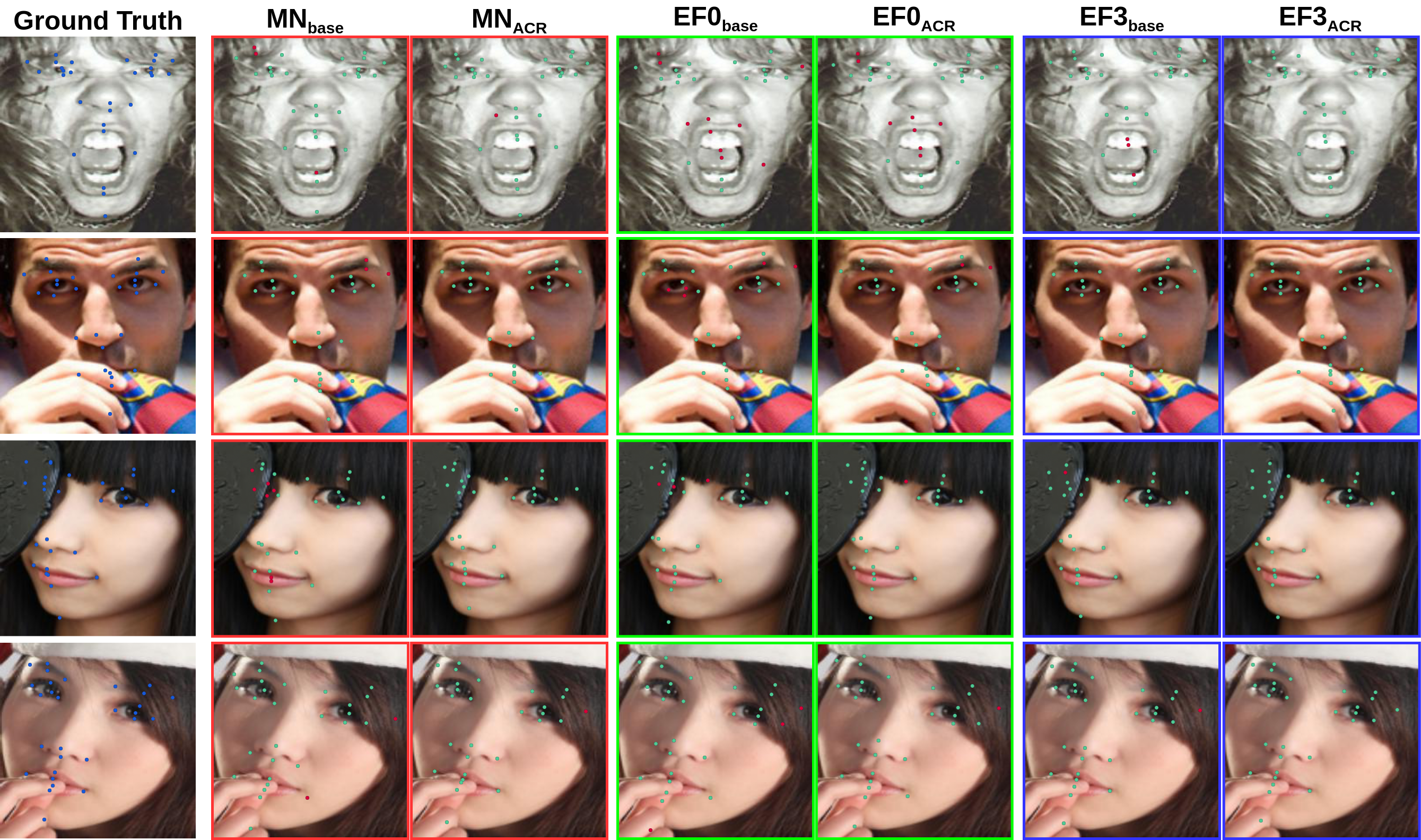}
  \caption{Examples of facial landmark points detection using our proposed ACR loss versus the L2 loss on COFW~\cite{burgos2013robust} dataset. For each landmark point, if the error rate is more than $0.1$, it is considered as a failure and we printed it as red, and otherwise it is green.}.
  \label{fig:ACR_cofw_samples}
\end{figure*}

\subsection{Qualitative Results Evaluation}\label{sec:qualitive_evaluation}
In order to investigate the qualitative performance of our proposed ACR loss, we provide some examples of facial landmark detection using MN\textsubscript{ACR}, EF-0\textsubscript{ACR}, EF-3\textsubscript{ACR}, and compared to the corresponding networks trained using L2 loss, MN\textsubscript{base}, EF-0\textsubscript{base}, EF-3\textsubscript{base}. As both Fig.~\ref{fig:ACR_300w_samples} (example of face alignment using on 300W~\cite{sagonas2013300} dataset), and Fig.~\ref{fig:ACR_cofw_samples} (examples of facial landmark detection on COFW~\cite{burgos2013robust} dataset) show, the qualitative performance of the models trained with ACR loss are much better than the corresponding models trained with L2 loss, indicating the effectiveness of our proposed ACR loss in improvement of CBR face alignment task.

\end{document}